\documentclass[letterpaper, 10 pt, conference]{ieeeconf}  

\IEEEoverridecommandlockouts                              

\usepackage{algorithm}
\usepackage{algpseudocode}
\usepackage{float}
\usepackage{hyperref}
\usepackage{graphicx}
\usepackage{xcolor}
\usepackage{booktabs}
\usepackage{amsmath}
\usepackage{balance}

\usepackage{amsmath,multirow}

\newboolean{include-notes}
\setboolean{include-notes}{true}
\newcommand{\rhnote}[1]{\ifthenelse{\boolean{include-notes}}%
 {\textcolor{blue}{\textbf{RH: #1}}}{}}
\newcommand{\todo}[1]{\ifthenelse{\boolean{include-notes}}%
 {\textcolor{red}{\textbf{TODO: #1}}}{}}

\hypersetup{
    hidelinks,
    pdftitle={I Won't Know Until I Try: Interactive Information Gathering for Long-Horizon Manipulation},
    pdfauthor={Linfei Yuan and Rachel Holladay}
}

\title{\LARGE \bf
I Won’t Know Until I Try: \\ Interactive Information Gathering for Long-Horizon Manipulation}

\author{Linfei Yuan and Rachel Holladay%
\thanks{\raggedright
Both authors are with the GRASP Laboratory,
University of Pennsylvania, Philadelphia, PA 19104, USA.
{\tt\small \{yuanlf,rhollada\}@engineering.upenn.edu}.}%
}

\begin{document}

\maketitle
\thispagestyle{empty}
\pagestyle{empty}


\begin{abstract}
Long-horizon manipulation often depends on object properties that cannot be reliably inferred from appearance alone. Physical interaction can reveal these properties, but it can also change the world in ways that invalidate subsequent actions. We present a task and motion planning (TAMP) framework that integrates task-directed interactive information-gathering with interleaved planning and execution. Building on PDDLStream, our framework plans optimistically with these interactive actions, verifies their outcomes during execution, and adapts the remaining plan by retaining still-applicable actions and replanning only the connections needed to continue execution. We evaluate the framework in simulation and on a real robot across tasks that require determining whether an object is movable, a pen has ink, a drawer can be opened, or a container is empty. These experiments demonstrate how interaction-driven state updates and recovery replanning support continued execution under uncertain outcomes and external disturbances.

\end{abstract}

\section{Introduction}
\label{sec:introduction}

\begin{figure}[t]
    \centering
    \includegraphics[width=0.8\columnwidth]{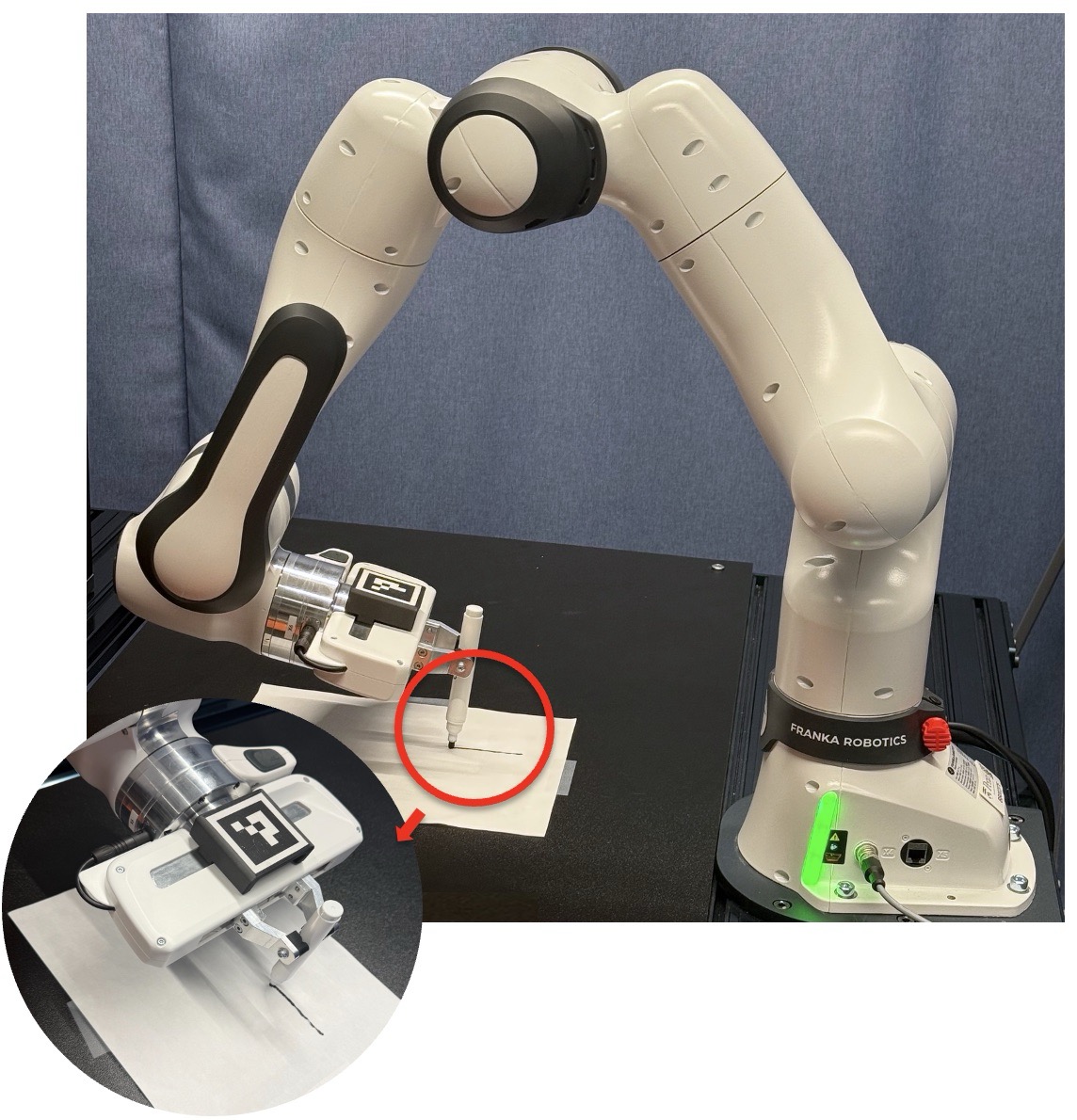}
    \caption{Testing pen functionality through physical interaction.
    The robot attempts to draw on paper and checks for ink marks to determine whether the pen is functional.}
    \label{fig:pen_draw_interaction}
\end{figure}

Long-horizon manipulation requires a robot to determine not only how
to act, but also what it needs to know before acting. 
While passive perception can often identify objects and estimate their poses, some task-relevant properties cannot be reliably observed from appearance alone, such as whether an object is movable, a drawer is unlocked, a non-transparent container is empty, or a pen has ink (Fig.~\ref{fig:pen_draw_interaction}).
To resolve this missing information, the robot must employ \textit{interactive perception}: using physical interactions and their observed outcomes to reveal these object properties~\cite{interactive_perception}.

For example, testing whether an object is movable may require pushing it slightly.
While the push reveals the object's movability,
 it may also displace the object by an uncertain offset.
Such an interaction can therefore both \textit{resolve uncertainty and alter the physical state} for subsequent actions.

Consequently, information gathering must be integrated into planning to determine which properties need to be revealed to accomplish the task while also accounting for the physical effects that require downstream adaptation.




Plan repair and adaptive execution provide established mechanisms for revising plans when observations and changes disrupt the initial plan~\cite{online_replanning,tamper}.
Adopting this perspective, we address how execution can recover when interactive observations contradict initial assumptions. 
Rather than replanning globally from scratch after every interaction, our approach identifies and retains valid plan suffixes and constructs minimal local connections only when needed; a process we refer to as \textit{recovery replanning}.

We frame this problem of multi-step manipulation under uncertainty as an instance of Task and Motion Planning (TAMP). TAMP addresses the action selection problem by jointly searching over discrete action choices and continuous parameters, such as grasps and trajectories, that satisfy geometric, kinematic, and physical constraints~\cite{tamp_review,pddlstream}. 
Building on the TAMP framework PDDLStream, we incorporate  interactive, information-gathering actions, execution monitoring, and recovery
replanning. 

Our framework models task-relevant properties as initially unknown, using optimistic interactive action models that are verified during execution.
Observations update the state, while execution monitoring identifies when interaction outcomes or physical side effects render planned actions inapplicable or infeasible---for example, when a pick cannot be completed because an object is immovable, or when a previously planned motion is no longer collision-free. Rather than discarding the entire remaining plan, our framework retains its reusable portions and uses recovery replanning to reconnect the current state to a valid plan suffix. This allows execution to adapt to interaction outcomes and state changes while preserving prior planning effort whenever possible.


Our work makes the following contributions: 
\begin{itemize}
     \item Augmented an existing TAMP framework with optimistic, task-directed, interactive information-gathering actions for manipulation under uncertainty.

    \item Formulated an interleaved planning and execution method that leverages recovery replanning.


    \item Evaluated across 4 interaction domains (whether an object is movable, a pen has ink, a drawer openable, and a container empty) in simulation and on a real robot.
    
\end{itemize}


\section{Related Work}
\label{sec:related-work}

Our work builds on TAMP under partial observability, task-directed information gathering through interaction, and adaptive execution and online replanning. We study settings where physical interactions both reveal task-relevant properties and alter the world, requiring state updates and recovery that reuses applicable portions of the original plan.

\subsection{Task and Motion Planning under Partial Observability} 
Task and motion planning (TAMP) frames multi-step manipulation as a joint search of the sequence of actions for the robot to execute and the constrained discrete and continuous values, such as grasps, paths, objects, etc., that parameterize those actions~\cite{tamp_review}. 
While it is common to assume a fully known environmental state, real-world manipulation often depends on properties or outcomes that can only be uncovered during execution. 

To address partial observability, prior approaches incorporate perception and state estimation into the planning and execution process. 
For example, belief-space TAMP frameworks reason over probability distributions in the context of hierarchical planning~\cite{belief_tamp} or via probabilistic abstractions of closed-loop controllers to enable information gathering and risk-aware decision-making~\cite{tampura}. 
Other methods infer models for unknown objects directly from RGB-D observations and estimated affordances~\cite{estimated_affordances}. 

While these works establish mechanisms for planning under uncertainty, our framework takes a complementary, execution-oriented approach. 
Rather than tracking continuous belief distributions, we maintain explicit predicates for property knowledge, dynamically adapting the plan as observations update both these property predicates and the geometric relationships. Our representation builds on established symbolic encodings of known and unknown facts~\cite{classical_replanning,seeing_believing}, rather than introducing a new epistemic formalism.

\subsection{Task-Directed Information Gathering through Interaction}

When visual information is insufficient to determine object properties or states, interactive perception uses physical actions to elicit informative
sensory observations~\cite{interactive_perception}.
Rather than performing unguided exploration or generating data to train perceptual models~\cite{property_learning}, active information-gathering in TAMP problems are explicitly directed by the task. 
In the context of partially observable
Markov decision processes (POMDPs), this  perspective guides the search to reduce task-relevant uncertainty rather than reducing uncertainty
indiscriminately. 
Similarly, the recent framework Seeing is Believing uses foundation models to
ground beliefs and plan information-gathering actions under
property uncertainty~\cite{seeing_believing}. 

These frameworks often assume a strict decoupling between the information-gathering actions and actions that interact with the world, thus assuming that perception leaves the physical world unaffected. 
In contrast, we explicitly account for 
interactions
that have \textit{coupled physical and informational outcomes}: a push can
reveal movability and alter object pose, while lifting a bottle can reveal whether it is empty and change the robot's grasp state. 
Consequently, our execution must
account for both uncovered information and changes to the physical state.  


\subsection{Adaptive Execution and Online Replanning}

When the execution of a plan deviates from expectation, due to unexpected physical dynamics or unobserved properties, the na\"ive approach is to re-generate an entirely new plan, i.e. to replan from scratch. 
To avoid this, planning repair seeks to reuse some components of the existing solution, ranging from the action sequence to the continuous trajectories to the closed-loop behaviors. 
Van der Krogt and de Weerdt formulated planning repair as removing obstructing actions or constraints and adding actions needed to restore a valid plan~\cite{plan_repair}. 

Approaches such as SS-Replan~\cite{online_replanning} initially constrain online replanning to reuse the unexecuted plan tail, preserving action sequencing, object selections and grasps while recomputing only observation-dependent poses, configurations, and trajectories. 
This constraint is relaxed if no suitable plan is found within a timeout. 

Alternatively, the action space of TAMPER~\cite{tamper} includes closed-loop actions that cannot be fully ground prior to execution.
Once executed, an online motion repair connects the resulting configuration to the next planned action. 

Other frameworks achieve adaptive execution by limiting the planning horizon~\cite{receding_horizon} or by expressing motions relative to objects~\cite{object_centric}. 
LESAMPLE~\cite{lesample} samples scene hypotheses from low-entropy perceptual beliefs and replans when observations invalidate the current plan, while Levihn et al.~\cite{foresight} use belief monitoring and selective hierarchical replanning. 
Robust logical-dynamical systems~\cite{rlds} govern behavior activation and continuation through logical conditions, supporting reactive plan execution.

Learning-based approaches adapt execution by updating failure probability models across tasks~\cite{failure_option} or training reinforcement learning bridge policies to handle state discrepancies~\cite{bridge_policy}. 
For the latter, the policy learns when to return control to the planner,
without requiring reconnection to a specified original-plan suffix.

Prior work provides complementary strategies for adapting execution
and reusing prior planning effort. 
Our framework retains validated, potentially reusable, instantiated action suffixes and only generates new actions when execution encounters a gap. 

\section{Problem Formulation}
\label{sec:background}

We define our problem using the Planning Domain Definition Language (PDDL)~\cite{mcdermott1998pddl}.  
In PDDL, a \emph{predicate} is a Boolean-valued expression capturing a relationship between state variables. 
Grounding a predicate's arguments produces a \emph{fact} and the state of the world $s$ is defined by a set of these facts.
For example, \texttt{(Mat m)} denotes that \texttt{m} as a mat, \texttt{(IsMovable o)} that object \texttt{o} can be moved, and \texttt{(On o m)} that object \texttt{o} is on mat \texttt{m}. 

Actions are defined as controllers parameterized by discrete and continuous variables.  An action $a_i$ is characterized by its \emph{preconditions}, $\operatorname{Pre}(a_i)$, the set of facts that must hold for the controller to execute, and its \emph{effects},
$\operatorname{Eff}(a_i)$, the set of facts true upon execution.
All of our domains share a basic set of actions (\texttt{move\_free}, \texttt{pick}, \texttt{move\_holding}, \texttt{place}), supplemented with domain-specific manipulation and information-gathering actions.

Given an initial state $s_0$, a action set $\mathcal{A}$, our objective is to find a plan $\pi=[a_1,\ldots,a_n]$ that achieve the goal state $G$, specified as a set of target facts. 
We cast this as a task and motion planning (TAMP) problem, where we must solve for the sequence of actions and their corresponding discrete and continuous parameters~\cite{tamp_review}. 
We specifically use PDDLStream, a state-of-the-art TAMP framework that reduces the hybrid discrete/continuous planning problem to a sequence of discrete planning problems via focused sampling of the continuous and discrete parameters~\cite{pddlstream}.
We write $s \models G$ if the state $s$ satisfies all of the goal facts $G$, and $s \not\models G$ otherwise.

In each of our settings, the initial state specifies object types, object poses and the robot configuration.  
Crucially, certain task-relevant binary properties, such as whether an object is movable or whether a pen has ink remaining, are initially unknown. 
We construct tasks such that knowing these hidden properties is essential to achieving the goal, such as cleaning up all movable objects or storing only pens that contain ink. 

To uncover these properties, we equip the robot with information-gathering actions that interact with the environment, such as pushing an object to test for movability or writing with a pen to test for ink. 
Executing these actions affects the state in two ways: it alters geometric relations, such as object poses, while simultaneously revealing property facts. 
Consequently, the planner must generate a plan that incorporates these information-gathering actions while adapting to newly discovered facts and changes to the geometric world.


\section{Interactive, Information-Gathering Actions}
\label{sec:interactive-belief}

We focus on domains where initially unknown task-critical properties must be discovered through interaction. 
In these settings, a missing positive fact is ambiguous. 
It could indicate that property has not yet been evaluated or that it has been evaluated and found to be false. 
These cases demand different responses. 
In the former, the robot must execute an information-gathering action while in the latter it must adapt the remaining plan rather than re-evaluating the same property. 
Therefore we must track both
\emph{whether the property has been evaluated} and \emph{the resulting outcome}.

For each binary property $X$, we introduce two predicates: \texttt{CheckedX(...)} indicates that an action has evaluated the property, and \texttt{IsX(...)} represents its positive outcome.
Together, they encode three information statuses:
\begin{center}
{\footnotesize
\begin{tabular}{lcc}
\hline
\textbf{Property status} & \texttt{CheckedX} & \texttt{IsX} \\
\hline
Unknown          & absent  & absent  \\
Tested positive  & present & present \\
Tested negative  & present & absent  \\
\hline
\end{tabular}
}
\end{center}

Thus, a negative result is explicitly represented by
\texttt{CheckedX(...)} without \texttt{IsX(...)}. 
The absence of \texttt{IsX(...)} alone is insufficient. This representation builds on symbolic encodings of
known and unknown facts~\cite{classical_replanning,seeing_believing}.
We treat every interaction evaluation as deterministic, assuming the outcome resolves
the queried property to the fidelity required by the task, without tracking confidence scores or probabilities.

To plan prior to interaction, before the property values are known, we model the information-gathering actions with \textit{optimistic effects}.
As illustrated below, the \texttt{interact-x} action asserts that the property has been evaluated (\texttt{CheckedX(...)}) and assumes a positive outcome (\texttt{IsX(...)}). 
These effects serve planning predictions, rather than assertions that an
untested property is already known.
At execution time, the robot verifies whether \texttt{IsX(...)} holds, as described in Section~\ref{sec:execution-framework}. 
A downstream manipulation action \texttt{dependent-action} has \texttt{IsX(...)} as a precondition:

\noindent
\begin{minipage}[t]{0.48\linewidth}
{\footnotesize
\begin{verbatim}
(:action interact-x
  :precondition (and
    ...
    (not (CheckedX ...)))
  :effect (and
    ...
    (CheckedX ...)
    (IsX ...)))
\end{verbatim}
}
\end{minipage}%
\hfill
\begin{minipage}[t]{0.48\linewidth}
{\footnotesize
\begin{verbatim}
(:action dependent-action
  :parameters (...)
  :precondition (and
    ...
    (IsX ...))
  :effect (and ...))
\end{verbatim}
}
\end{minipage}
\vspace{0.2cm}

When $X$ is initially unknown, the planner naturally sequences the test
before the dependent action. Information gathering is therefore
arises directly from the action's preconditions rather than
as an explicit preprocessing stage.
The \texttt{CheckedX} precondition prevents the planner from repeatedly executing the same test after a negative outcome.

As a concrete example, consider our aforementioned task where the robot must clean up all movable objects by placing them on a mat. Here, \texttt{CheckedMovable(o)} and \texttt{IsMovable(o)} track the evaluate status and outcome, with \texttt{push} acting as the information-gathering action.
Because downstream actions like \texttt{pick}, \texttt{move\_holding}, and \texttt{place} require \texttt{IsMovable(o)}, the planner inserts a \texttt{push} action before interacting with the untested object.

The following section describes how these property updates, together with observations of the physical state, guide execution and adaptation of the remaining plan.

\section{Interleaved Planning and Execution}
\label{sec:execution-framework}
\label{sec:belief-update}

While the optimistic effects facilitate an initial plan, execution
may reveal different property values or lead to unexpected states. 
An information-gathering action may shift an object to an unexpected pose, fail to move it or lead the robot in an unexpected configuration. 
We therefore interleave execution with state updates, monitoring, and
recovery replanning.

\begin{figure*}[t]
    \centering
    \includegraphics[width=\textwidth]{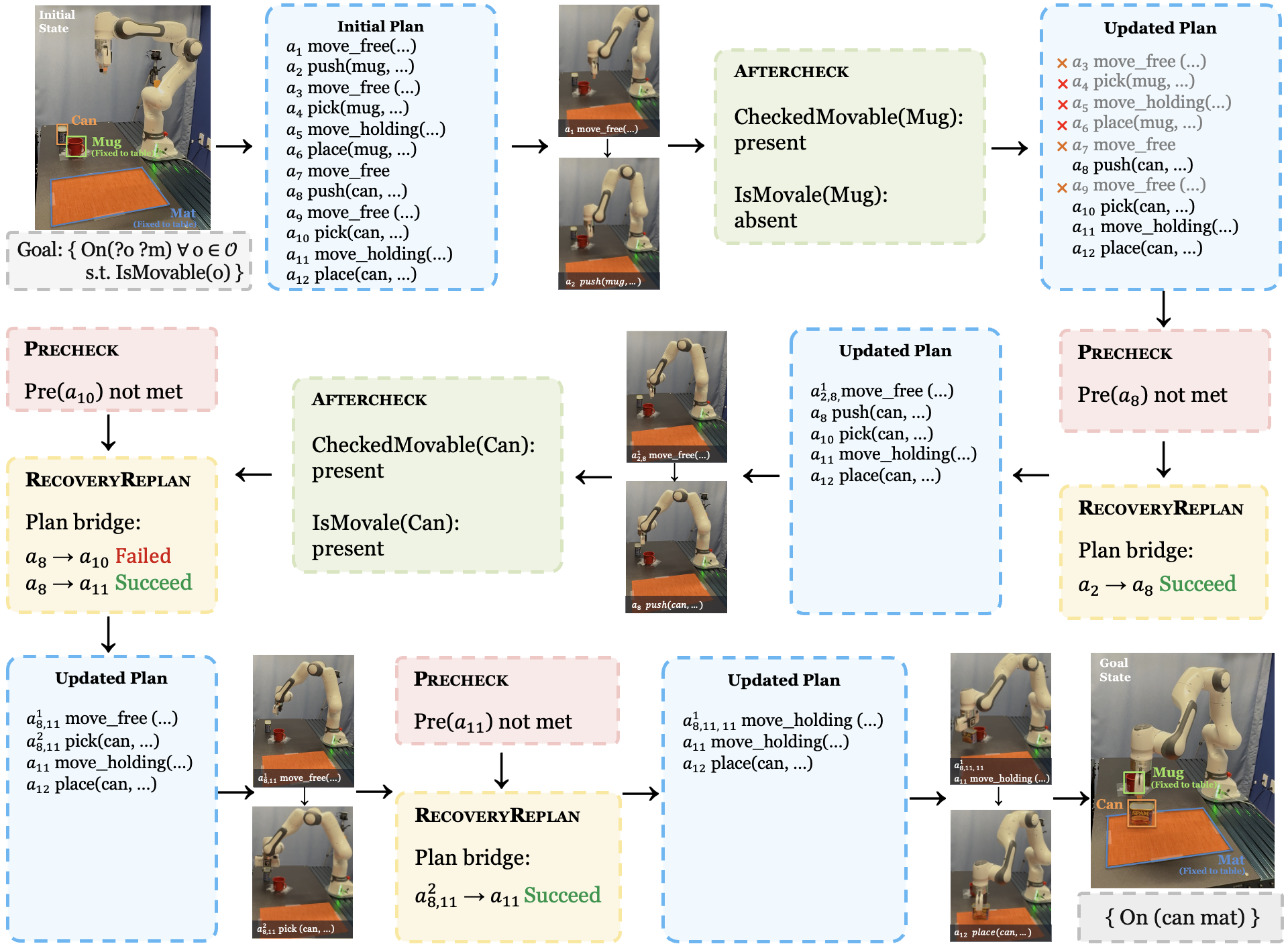}
   \caption{Example execution trace with online monitoring and recovery.
Blue blocks show the initial plan and successive updated plans, containing retained actions and newly introduced bridging actions.
The initial plan consists of actions $a_1,\ldots,a_{12}$, while $a^k_{i,j}$ denotes the $k$th action of a bridge plan reconnecting execution after $a_i$ to retained action $a_j$.
After each interactive information-gathering action, \textsc{Aftercheck} updates the symbolic state from the observed outcome (green blocks), recording that the object has been checked and whether the tested property holds.
\textsc{Validator} removes actions that are no longer valid: orange crosses denote collision-infeasible actions, while red crosses denote actions whose symbolic preconditions are infeasible.
Before each retained action, \textsc{Precheck} verifies executability (red blocks).
Detected precondition mismatches trigger \textsc{RecoveryReplan} (yellow blocks), which searches for a bridge to the next retained action or, if unsuccessful, to a later retained action.
Each successful repair produces an updated plan combining the recovered bridge with the retained suffix.
The system executes the bridge and resumes the retained plan, repeating state updates, validation, and on-demand repair until the goal is reached.
Robot snapshots show the corresponding physical execution from the initial scene to the goal state.}

    \label{fig:execution_trace}
\end{figure*}

Algorithm~\ref{alg:execution} details our planning and execution framework and Figure~\ref{fig:execution_trace} provides an illustrative example. 
Here the goal is to place all movable objects on the orange mat, where it is initially unknown if the red mug or can are movable.

We generate an initial plan by calling PDDLStream~\cite{pddlstream} (line 2). 
If successful, the robot enters the execution loop.

Each iteration starts with \textsc{ObserveState} (line~7), updating the object poses and robot configuration,  while retaining previously learned property facts. 
This update can capture action outcomes, external disturbances and execution failures. If this state satisfies $G$, execution successfully terminates (lines~8--10). 

\begin{algorithm}[t]
\caption{Interleaved Planning and Monitored Execution}
\label{alg:execution}
\footnotesize
\begin{algorithmic}[1]
\Require Current planning state $s$, active goal set $G$

\While{$s \not\models G$}
    \State $\pi \gets \Call{Plan}{s,G}$
    \If{$\pi = \textsc{Failure}$}
        \State \Return failure
    \EndIf

    \While{\textsc{True}}
        \State $s \gets \Call{ObserveState}{s}$
        \If{$s \models G$}
            \State \Return success
        \EndIf

        \State $\pi_{\mathrm{skel}} \gets
            \Call{Validator}{\pi,s}$
        \If{$\pi_{\mathrm{skel}} = [\,]$}
            \State \textbf{break}
                \Comment{global planning}
        \EndIf

        \State $a \gets
            \Call{FirstAction}{\pi_{\mathrm{skel}}}$

        \If{\Call{Precheck}{$a,s$} $= \textsc{False}$}
            \State $(\rho,a_i) \gets
                \Call{RecoveryReplan}{s,\pi_{\mathrm{skel}}}$

            \If{$\rho = \textsc{Failure}$}
                \State \textbf{break}
                    \Comment{global planning}
            \EndIf

            \State $\pi \gets \rho \mathbin{\Vert}
                \Call{SuffixFrom}{\pi_{\mathrm{skel}},a_i}$
            \State \textbf{continue}
                \Comment{revalidate the updated plan}
        \EndIf

        \State $s^{-} \gets s$
        \State \Call{Execute}{$a$}
        \If{\Call{IsInteractive}{$a$}}
            \State $s^{+} \gets \Call{ObserveState}{s}$
            \State $(s,G) \gets
                \Call{Aftercheck}{s^{-},s^{+},G,a}$
        \EndIf

        \State $\pi \gets
            \Call{SuffixAfter}{\pi_{\mathrm{skel}},a}$
    \EndWhile
\EndWhile

\State \Return success
\end{algorithmic}
\end{algorithm}

Otherwise, execution monitoring evaluates the remaining plan with the updated state by separating two questions: which actions are still valid and which action can be executed now. 
To address the first, \textsc{Validator} forward propagates the state through the  remaining plan, removing any actions who's preconditions are no longer valid (Line~11). 
For example, in Fig.~\ref{fig:execution_trace}, the mug's revealed immovability invalidates the subsequent \texttt{pick}, \texttt{move\_holding}, \texttt{place} actions (red cross marks). 
Additionally, its current pose is in collision with previously generated trajectories, thus invalidating three \texttt{move\_free} trajectories (orange cross marks). 

The resulting plan $\pi_{\mathrm{skel}}$ may contain gaps, such as missing transition motions, where now invalid actions were removed. 
Rather than eagerly repairing the gaps, we lazily defer to avoid premature planning under uncertain conditions. 
If no usable actions remain, we replan from scratch (lines~12--14).

Otherwise, \textsc{FirstAction} selects the next retained action and
\textsc{Precheck} evaluates its preconditions (lines~15--16). 
Small perceptual discrepancies are handled via state matching tolerances.
For example, an observed configuration $q_{\mathrm{obs}}$ matches the action's starting configuration $q_{a}$ when $||q_{\mathrm{obs}}-q_{a}\|_2 \leq \delta_q$ for some $\delta_q$ (with analogous checks object poses). 
In Fig.~\ref{fig:execution_trace}, the next \texttt{push(can)} remains in $\pi_{skel}$ but fails its precheck due to configuration mismatch
This illustrates that action reusability does not imply immediate action readiness.

\begin{figure}[t]
    \centering
    \includegraphics[width=0.5\textwidth]{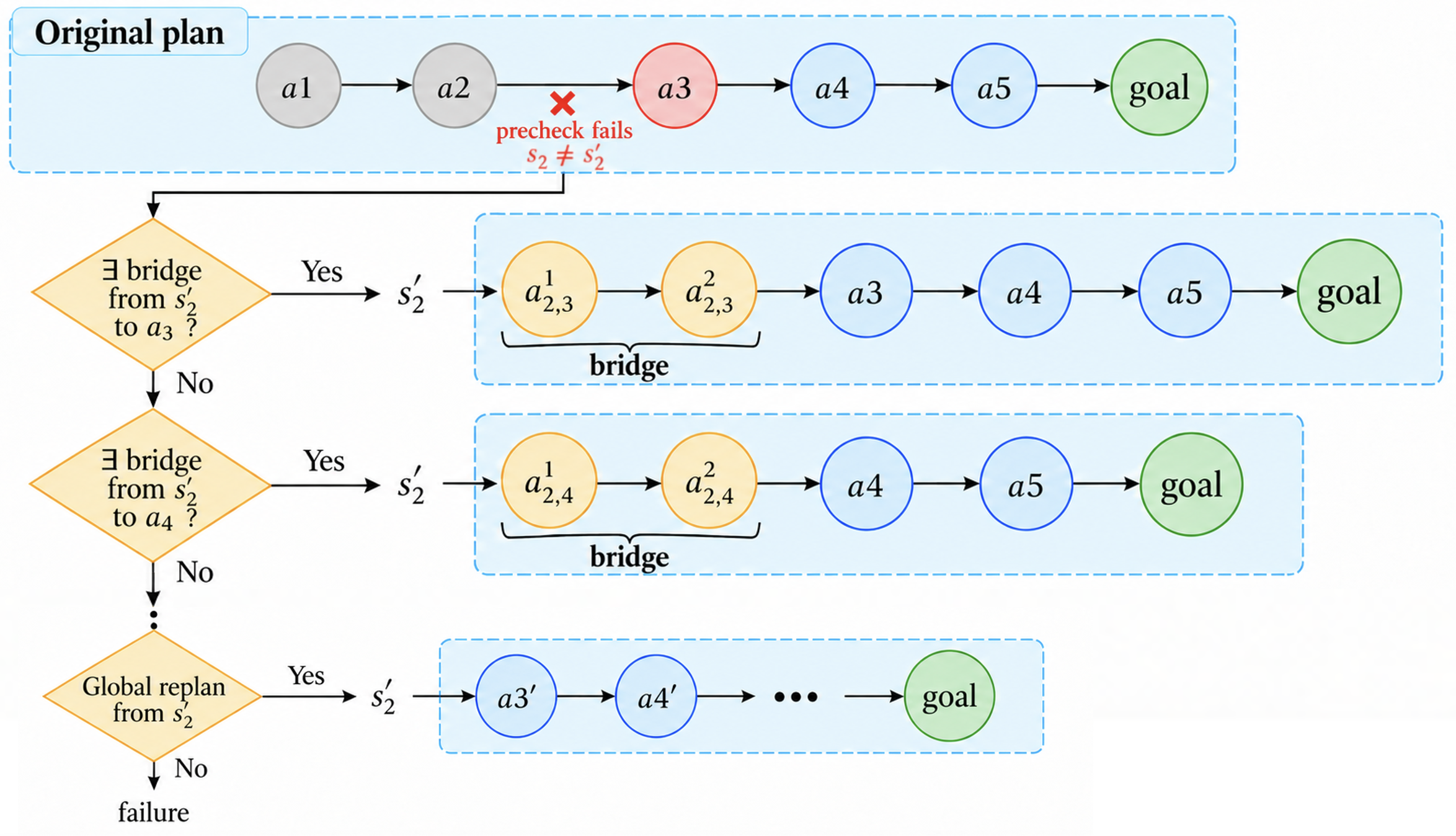}
    \caption{Recovery replanning through bridge construction.
After executing $a_1$ and $a_2$ (gray), the current planning state $s_2'$ differs
from the predicted state $s_2$, and $a_3$ fails its \textsc{Precheck} (red).
\textsc{RecoveryReplan} searches from $s_2'$ for a bridge that establishes
$\operatorname{Pre}(a_3)$; if unsuccessful, it tries later retained actions in order. 
The first bridge found is prepended to the retained suffix beginning at
its target action. If no bridge is
found to any retained action, global replanning searches from
$s_2'$ toward the active goal set $G$ (green), producing a new
action sequence $[a_3',a_4',\ldots]$. If this search also fails,
the algorithm returns failure.
    }
    \label{fig:recovery_replan_figure}
\end{figure}

\label{sec:plan-repair}
When a \textsc{Precheck} fails, \textsc{RecoveryReplan} calls PDDLStream to find a sequence of actions $\rho$ that \emph{bridge} from the current state
$s$ to a goal state defined as the preconditions of a downstream action $a_{i} \in \pi_{skel}$. 
Candidates are evaluated sequentially starting from the next action, favoring earlier plan re-entry. 
Figure~\ref{fig:recovery_replan_figure} illustrates this search from
the observed state $s_2'$: recovery first targets
$\operatorname{Pre}(a_3)$, falls back on $\operatorname{Pre}(a_4)$ if unsuccessful, and continues through later actions.
If no candidate lead to a successful bridge sequence, we plan from scratch (lines~18--20). 


When recovery succeeds, the bridge actions $\rho$ are prepended to the remaining plan suffix (line~21):
\[
    \pi \gets \rho \mathbin{\Vert}
    \operatorname{SuffixFrom}(\pi_{\mathrm{skel}},a_i),
\]
where $\Vert$ denotes sequence concatenation. Next, the algorithm loops back to observation and
validation (line~22). 
In Fig.~\ref{fig:execution_trace}, a \texttt{move\_free} bridge is
inserted before \texttt{push(can)} to account for the configuration mismatch.


Once prechecked, \textsc{Execute} performs the action (line~25).
For information-gathering actions,  the post-action state $s^{+}$ is observed, and
\textsc{Aftercheck} uses the change from $s^{-}$ to $s^{+}$ to update the property information by recording \texttt{CheckedX(...)} and adding or removing \texttt{IsX(...)} (lines~26--29).
In Fig.~\ref{fig:execution_trace} both mug and can are marked checked, yielding negative and positive movability results, respectively.
Additionally, \textsc{Aftercheck} incorporates the newly revealed property information to update the goals $G$.
In our example, the goal set is revised such that the immovable mug is removed from the target set but the movable can remains.

Removing the executed action from the plan (line~30), the entire loop repeats. 
Execution halts when $G$ is satisfied. If $\pi_{skel}$ is exhausted before achieving $G$, PDDLStream generates a new plan. 

\section{Experiments}
\label{sec:experiments}



We introduce four domains that require different types of  interactive information gathering. We primarily evaluate the efficiency of our framework as comparing to a global replanning baseline. Finally, we demonstrate our framework ability to handle unexpected execution disturbances and multiple interdependent information-gathering actions.


\subsection{Domains}
We define four manipulation domains:

\begin{figure}[t]
    \centering
    \begin{tabular}{cc}
        \includegraphics[width=0.48\columnwidth]{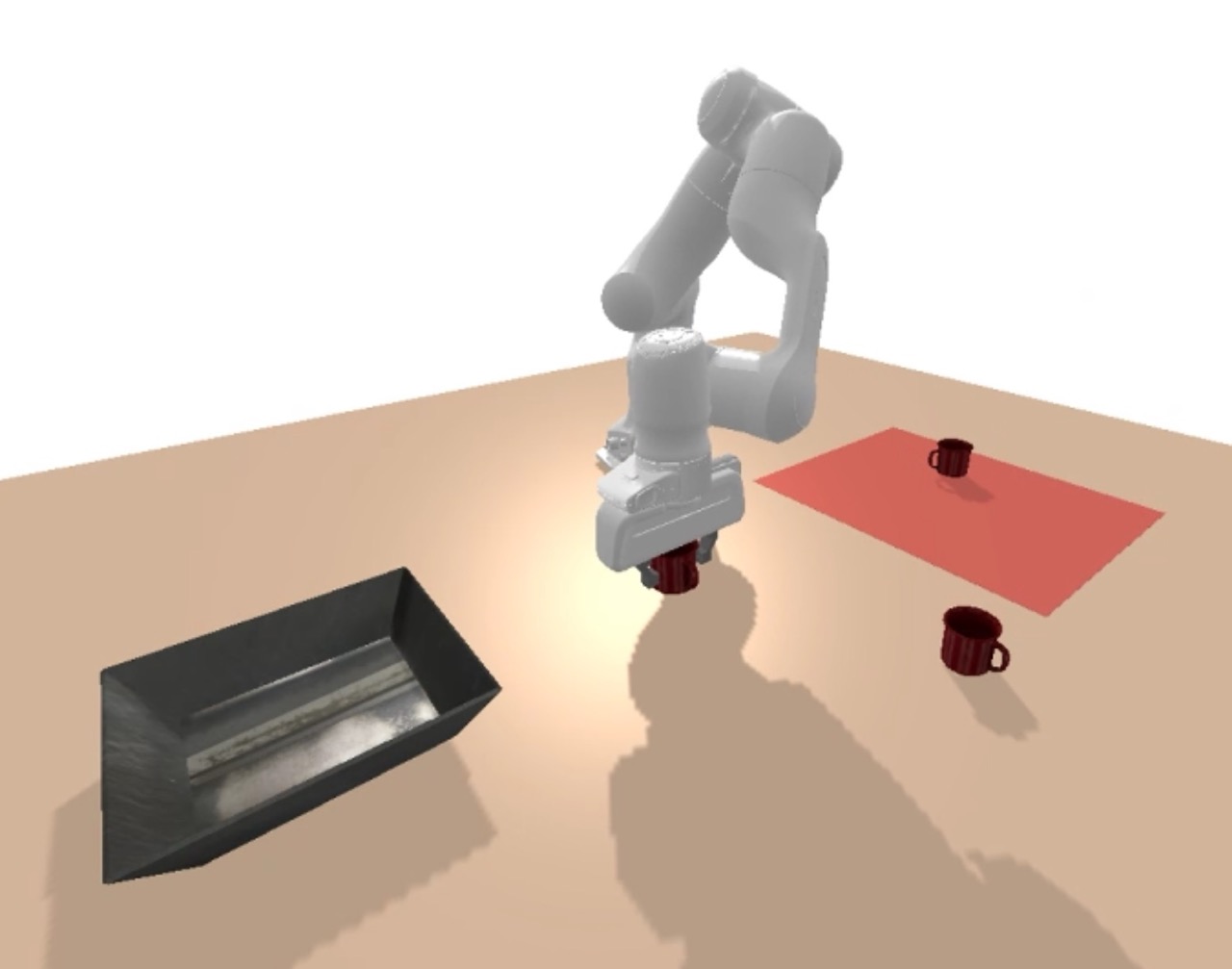} &
        \includegraphics[width=0.48\columnwidth]{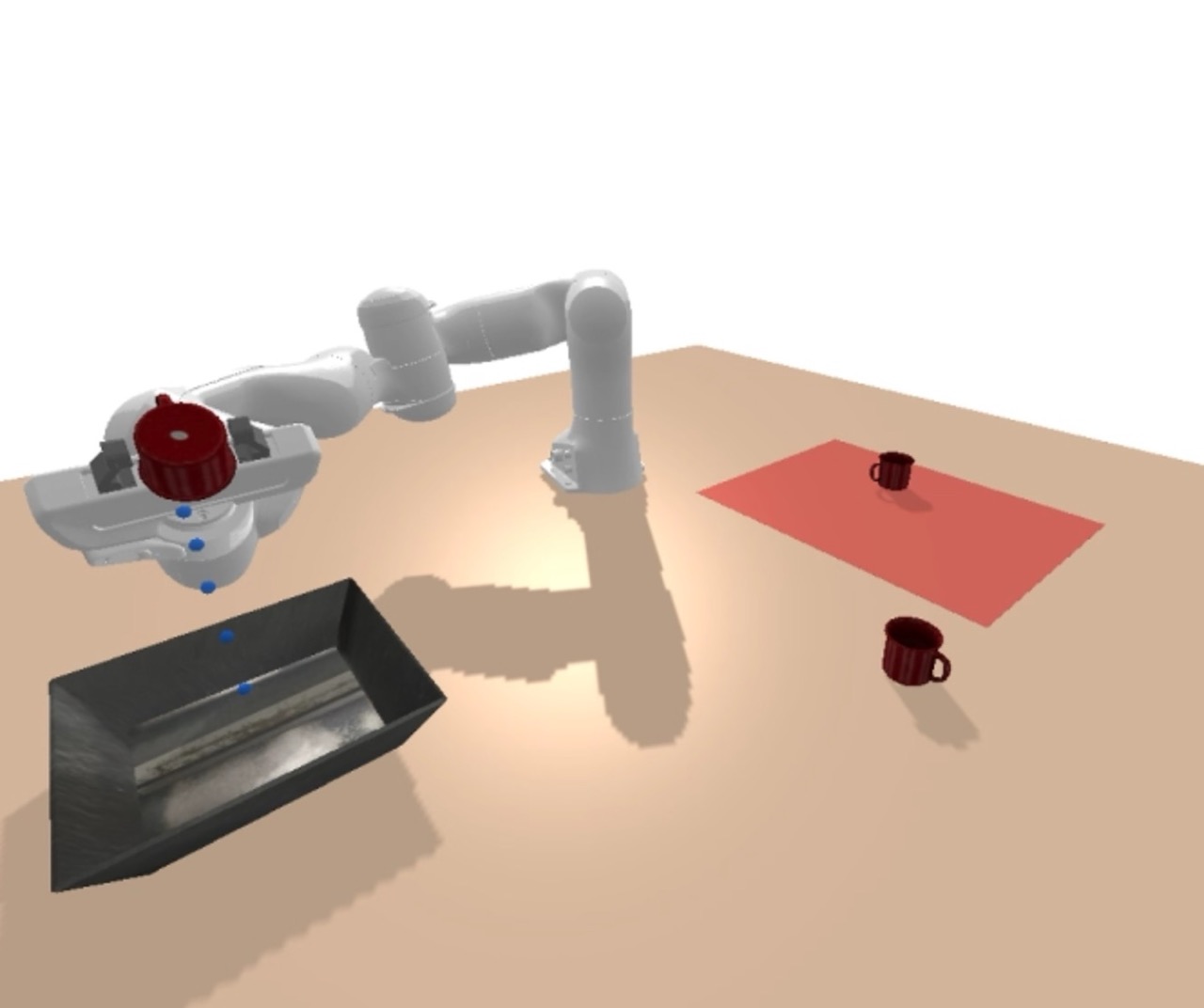} \\[2pt]
        \includegraphics[width=0.48\columnwidth]{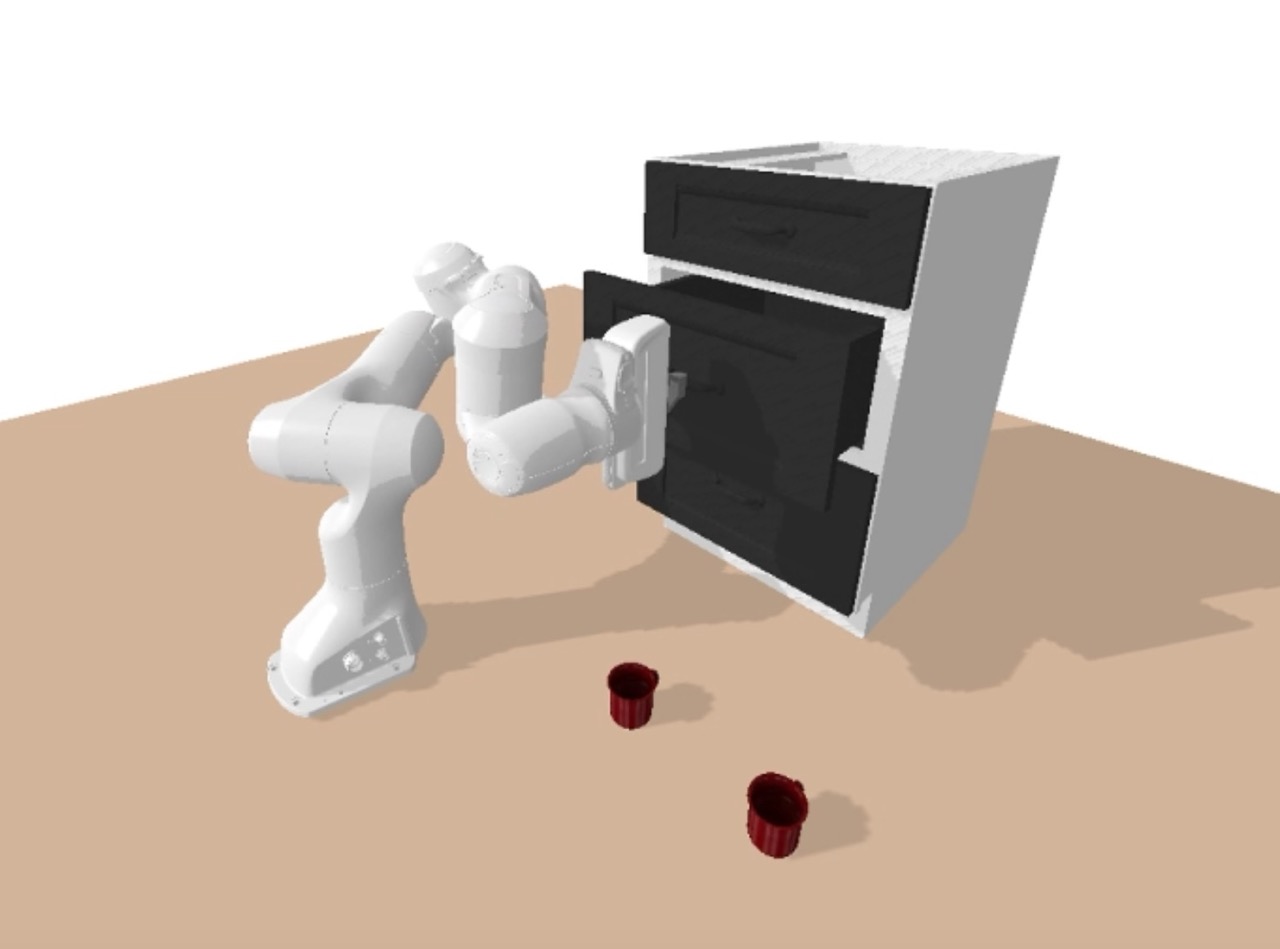} &
        \includegraphics[width=0.48\columnwidth]{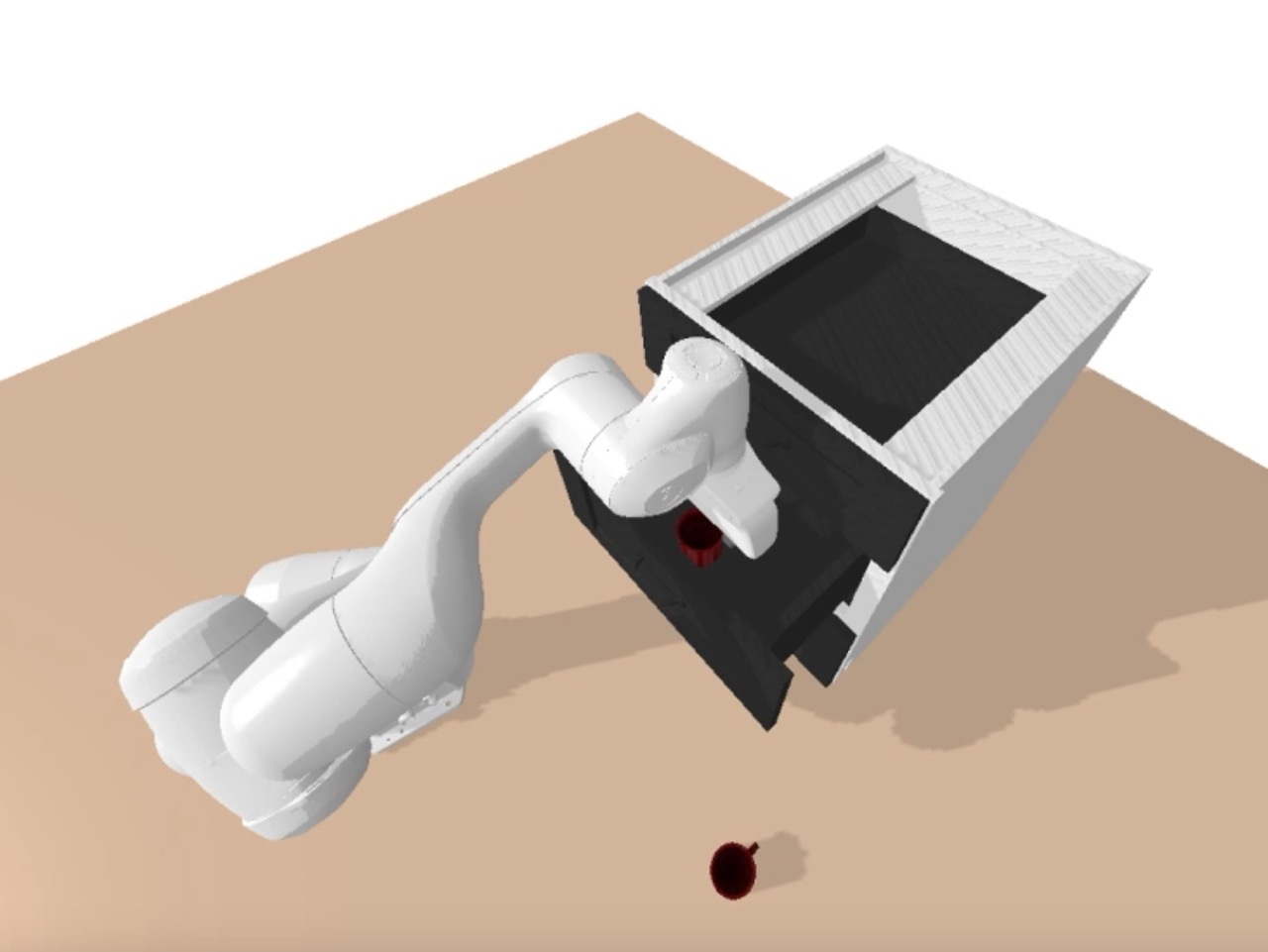}
    \end{tabular}
    \caption{Representative simulation experiments in two domains.
    Top: Container-Emptiness, where a negative interaction result
    (non-empty mug) leads from picking to pouring.
    Bottom: Drawer-Openability, where a positive interaction result
    (openable drawer) leads from pulling to placing the mug inside.}
    \label{fig:domain_overview}
\end{figure}

\noindent\textbf{Object-Movability:}
Sorts movable objects onto a target mat. The robot executes a probing push to observe displacement. Displaced and hence movable objects are transferred to the mat, while static objects remain in place.

\noindent\textbf{Pen-Functionality:}
Transfers functional pens to a target mat. The robot executes a test stroke on paper to inspect for ink. Functional pens are moved to the mat, whereas nonfunctional pens are returned to the workspace.

\noindent\textbf{Drawer-Openability:}
Stores objects in assigned drawers. The robot pulls each drawer handle to test articulation. Upon success, the object is placed inside, whereas objects assigned to locked drawers are left on the table (Figure~\ref{fig:domain_overview}).


\noindent \textbf{Container-Emptiness:}
Places empty containers onto a target mat. During pickup, the robot compares the measured mass against the expected mass. 
Empty containers are placed directly on the mat, while non-empty containers are first emptied via pouring (Figure~\ref{fig:domain_overview}).

We evaluate our framework in PyBullet~\cite{coumans2021pybullet}, using CPU-only computation. 
Each domain has two difficulty levels based on the number of targets: an \textit{easy} setting (2–3 objects) and a \textit{hard} setting (4–6 objects).
The identities of positive and negative targets vary across trials. 

\subsection{Effectiveness and Efficiency of Recovery Replanning}
\label{sec:exp-recovery}

\begin{table}[t]
\centering
\caption{Task completion rates over 20 trials}
\label{tab:completion}
\footnotesize
\setlength{\tabcolsep}{4pt}
\renewcommand{\arraystretch}{1.12}

\begin{tabular*}{\columnwidth}
{@{\extracolsep{\fill}}llccc@{}}
\hline
Domain & Level & Ours & Global & Joint $n$ \\[2pt]
\hline
Object-Movability   & Easy & 18/20 & 19/20 & 17 \\
   & Hard & 20/20 & 18/20 & 18 \\
Pen-Functionality    & Easy & 20/20 & 20/20 & 20 \\
 & Hard & 19/20 & 20/20 & 19 \\
Drawer-Openability & Easy & 19/20 & 20/20 & 19 \\
& Hard & 17/20 & 16/20 & 13 \\
Container-Emptiness   & Easy & 20/20 & 19/20 & 19 \\
 & Hard & 19/20 & 13/20 & 12 \\
\hline
\end{tabular*}

\end{table}

\begin{table*}[t]
\centering
\caption{SIMULATION EFFICIENCY RESULTS ACROSS 20-TRIAL}
\label{tab:main_results}
\footnotesize
\setlength{\tabcolsep}{2pt}
\renewcommand{\arraystretch}{1.15}
\setlength{\medmuskip}{2mu}

\begin{tabular*}{\textwidth}{@{\extracolsep{\fill}}lllccrrrrrr@{}}
\hline
\noalign{\vskip 2pt}
Domain & Level & \shortstack[l]{Object\\condition}
& Method & $n$
& $T_{\mathrm{off}}$
& $T_{\mathrm{on}}$
& $T_{\mathrm{exec}}$
& $T_{\mathrm{total}}$
& $L_0$
& $L_{\mathrm{exec}}$ \\[2pt]
\hline

\multirow{4}{*}{\shortstack[l]{Object-Movability}}
& \multirow{2}{*}{Easy}
& \multirow{2}{*}{\shortstack[l]{2 movable\\1 immovable}}
& Ours & 17
& $135.91 \pm 213.31$
& $42.51 \pm 49.32$
& $88.12 \pm 11.14$
& $\textbf{271.63} \pm 210.60$
& $19.47 \pm 1.94$
& $21.65 \pm 3.64$ \\

& & & Global & 17
& $90.50 \pm 102.41$
& $138.63 \pm 162.85$
& $85.50 \pm 8.91$
& $319.25 \pm 176.95$
& $19.06 \pm 1.48$
& $17.00 \pm 1.90$ \\

\cline{2-11}

& \multirow{2}{*}{Hard}
& \multirow{2}{*}{\shortstack[l]{3 movable\\2 immovable}}
& Ours & 18
& $274.78 \pm 199.55$
& $53.68 \pm 46.93$
& $165.02 \pm 13.95$
& $\textbf{507.94} \pm 205.49$
& $33.50 \pm 3.88$
& $33.83 \pm 4.05$ \\

& & & Global & 18
& $208.01 \pm 179.15$
& $527.23 \pm 376.16$
& $150.89 \pm 18.97$
& $902.73 \pm 357.15$
& $33.89 \pm 3.58$
& $29.56 \pm 3.47$ \\

\hline

\multirow{4}{*}{\shortstack[l]{Pen-Functionality}}
& \multirow{2}{*}{Easy}
& \multirow{2}{*}{\shortstack[l]{2 with ink\\1 without ink}}
& Ours & 20
& $28.11 \pm 6.99$
& $3.12 \pm 1.10$
& $17.84 \pm 1.96$
& $\textbf{58.29} \pm 8.66$
& $18.15 \pm 0.67$
& $19.10 \pm 0.45$ \\

& & & Global & 20
& $26.67 \pm 4.93$
& $8.62 \pm 6.10$
& $16.75 \pm 1.46$
& $61.53 \pm 9.30$
& $18.00 \pm 0.00$
& $19.00 \pm 0.00$ \\

\cline{2-11}

& \multirow{2}{*}{Hard}
& \multirow{2}{*}{\shortstack[l]{3 with ink\\3 without ink}}
& Ours & 19
& $77.77 \pm 18.08$
& $13.54 \pm 1.77$
& $35.80 \pm 2.58$
& $\textbf{177.33} \pm 18.58$
& $36.32 \pm 0.95$
& $39.21 \pm 0.63$ \\

& & & Global & 19
& $67.87 \pm 17.51$
& $99.38 \pm 36.28$
& $37.25 \pm 2.52$
& $270.99 \pm 49.86$
& $36.58 \pm 1.92$
& $39.00 \pm 0.00$ \\

\hline

\multirow{4}{*}{\shortstack[l]{Drawer-Openability}}
& \multirow{2}{*}{Easy}
& \multirow{2}{*}{\shortstack[l]{1 openable\\1 locked}}
& Ours & 19
& $124.36 \pm 140.62$
& $26.40 \pm 24.18$
& $26.54 \pm 8.65$
& $181.30 \pm 149.84$
& $16.42 \pm 2.34$
& $14.95 \pm 6.44$ \\

& & & Global & 19
& $100.35 \pm 86.40$
& $19.52 \pm 20.30$
& $22.32 \pm 5.11$
& $\textbf{145.85} \pm 92.23$
& $16.11 \pm 3.28$
& $10.95 \pm 4.44$ \\

\cline{2-11}

& \multirow{2}{*}{Hard}
& \multirow{2}{*}{\shortstack[l]{2 openable\\2 locked}}
& Ours & 13
& $249.26 \pm 209.06$
& $100.07 \pm 75.45$
& $46.98 \pm 5.44$
& $\textbf{409.97} \pm 236.38$
& $31.46 \pm 4.70$
& $23.08 \pm 4.84$ \\

& & & Global & 13
& $248.44 \pm 183.45$
& $235.25 \pm 143.42$
& $46.25 \pm 4.97$
& $548.99 \pm 257.61$
& $34.38 \pm 5.01$
& $20.15 \pm 2.88$ \\

\hline

\multirow{4}{*}{\shortstack[l]{Container-Emptiness}}
& \multirow{2}{*}{Easy}
& \multirow{2}{*}{\shortstack[l]{1 empty\\2 non-empty}}
& Ours & 19
& $21.14 \pm 10.74$
& $10.72 \pm 0.48$
& $11.43 \pm 2.13$
& $\textbf{46.40} \pm 10.06$
& $12.00 \pm 0.00$
& $18.00 \pm 0.00$ \\

& & & Global & 19
& $19.35 \pm 8.07$
& $23.28 \pm 11.59$
& $13.44 \pm 1.02$
& $60.37 \pm 11.05$
& $12.00 \pm 0.00$
& $18.32 \pm 0.95$ \\

\cline{2-11}

& \multirow{2}{*}{Hard}
& \multirow{2}{*}{\shortstack[l]{2 empty\\4 non-empty}}
& Ours & 12
& $65.25 \pm 25.65$
& $22.39 \pm 1.55$
& $26.12 \pm 2.35$
& $\textbf{128.26} \pm 26.75$
& $24.00 \pm 0.00$
& $36.08 \pm 0.29$ \\

& & & Global & 12
& $78.08 \pm 39.08$
& $237.47 \pm 113.44$
& $28.86 \pm 3.37$
& $367.49 \pm 106.17$
& $24.00 \pm 0.00$
& $37.67 \pm 2.06$ \\

\hline
\end{tabular*}

\par\smallskip
\begin{minipage}{\textwidth}
\footnotesize
\raggedright
Times are in seconds; values are mean $\pm$ sample standard deviation
over all jointly successful pairs ($n$) within each 20-trial.
\end{minipage}
\end{table*}

We first evaluate our framework's ability to efficiently adapt and replan compared to a \emph{Global} replanning baseline, which replans from scratch upon each state mismatch. Although SS-Replan~\cite{online_replanning} and TAMPER~\cite{tamper} also exploit plan reuse, a matched comparison would require adapting their planning and execution models to incorporate our interaction tests, property updates, and goal-revision rules.
We therefore use \emph{Global} as a controlled ablation to isolate the contribution of reconnecting to retained actions through bridge construction.
Both our framework and the baseline share the same action sets, state monitoring and updates and goal updates.


Across four domains and two difficulty levels, we evaluate 20 trials per condition (160 total trials per method). 
Global replanning attempts are capped at $400\,\mathrm{s}$ per query (up to three attempts or $900\,\mathrm{s}$ total), local recovery planning is capped at $120\,\mathrm{s}$, and maximum overall trial duration is set to $7,200\,\mathrm{s}$.

We evaluate our framework across three dimensions: task completion rate, planning and execution efficiency and recovery effectiveness. 
Tables~\ref{tab:completion}--\ref{tab:recovery_fallback} reports each of these metrics across domains and difficulty levels. 

With respect to task completion both methods achieve high success in the Easy cases but our method outperforms the Global baseline for Hard cases. 


To evaluate efficiency, we condition on task success and jointly analyze successful trial pairs ($n$), where both methods succeeded in the same randomized environment. 
Table~\ref{tab:main_results} details the 
initial offline planning time $T_{\mathrm{off}}$, 
the accumulated online planning time (including all recovery and global planning calls) $T_{\mathrm{on}}$, 
accumulated action-execution time $T_{\mathrm{exec}}$, 
end-to-end completion time $T_{\mathrm{total}}$, 
the initial plan length $L_0$ and the number of executed actions
$L_{\mathrm{exec}}$  as mean $\pm$ standard deviation.

Our framework achieves a lower completion time $T_{\mathrm{total}}$  in 7 of 8 settings, with the largest gains in the harder domains. In particular, compared with Global,
$T_{\mathrm{total}}$ is reduced by approximately $44\%$, $35\%$, and
$65\%$ for Object-Movability (Hard), Pen-Functionality (Hard), Container-Emptiness (Hard), respectively. Drawer-Openability (Easy) is the only case in which Global is faster.

This efficiency primarily stems from a reduction in online planning time $T_{\mathrm{on}}$, which drops by up to $91\%$ because generating shorter bridge actions is far less computationally expensive than replanning from scratch. 
These computational savings outweigh execution costs. 
Even when our framework requires addition actions, the overall runtime remains lower. 


Finally, regarding recovery effectiveness, Table~\ref{tab:recovery_fallback} shows that local recovery consistently reconnects to the remaining plan without invoking global replanning. 
Across the 152 successful trials with our framework, only 14 recovery-to-global fallback events occurred, with zero fallbacks required for domains Pen-Functionality or Container-Emptiness
Fallbacks primarily occur when a short remaining plan suffix provides a few viable reconnection points. 
Overall, the low fallback frequency
demonstrates that the local recovery effectively resolves most execution
deviations without incurring full replanning overhead.

\begin{table}[t]
\centering
\caption{Total recovery-to-global fallback events in Ours}
\label{tab:recovery_fallback}
\footnotesize
\setlength{\tabcolsep}{3pt}
\renewcommand{\arraystretch}{1.12}

\begin{tabular*}{\columnwidth}
{@{\extracolsep{\fill}}lcccc@{}}
\hline
\noalign{\vskip 2pt}
Level
& \shortstack{Object-\\Movability}
& \shortstack{Pen-\\Functionality}
& \shortstack{Drawer-\\Openability}
& \shortstack{Container-\\Emptiness} \\[2pt]
\hline
Easy & 5/18 & 0/20 & 1/19 & 0/20 \\
Hard & 4/20 & 0/19 & 4/17 & 0/19 \\
\hline
\end{tabular*}
\end{table}

Our interleaved planning and execution framework is inherently robust to unexpected disturbances or execution outcomes.
To demonstrate this, we evaluate the Container-Emptiness task (easy setting, 10 trials) under two unmodeled edge cases: (1) \textit{external scene changes}, where an object is mid-execution relocated to a random position to simulate intervention by another agent, and (2) \textit{action failures}, where the robot fails to pick up an object some fraction of the time. 
Without any modifications, our execution-monitoring and replanning loop automatically detects the resulting state mismatches and recovers to the goal, achieving 100\% (10/10) completion rate across both conditions.

\subsection{Transfer across Interaction Types and Task Composition}
\label{sec:exp-composition}

To evaluate our framework's ability to coordinate multiple, interdependent information-gathering actions, we crafted a combined domain that integrates all previously described interactions. 
The environment contains
two movable mugs (one empty, one non-empty), and two pens (one functional, one non-functional) with initially hidden properties. The goal is to store all of the functional pens and empty, movable mugs inside a drawer. 
The scenario requires the planner to simultaneously reason over movability tests, emptiness checks, writing tests, and drawer interaction in a single planning horizon. 
In this combined domain, our framework achieves 5 successful trials out of 5. 

\subsection{Real-Robot Demonstrations}
\label{sec:exp-real}

We demonstrate the Object-Movability and Pen-Functionality tasks on the Franka Emika FR3~\cite{haddadin2024franka}. 
Two Realsense D435i cameras provide 6D pose estimates via FoundationPose~\cite{foundationpose}, using given object meshes. 
Semantic category labels are supplied separately to the framework and we assume proprioception for the robot state. 
We evaluate the outcome of the pen's test stroke by thresholding and comparing before and after images of the writing surface. 
Most action controllers are implemented with position control. 
The \texttt{push} action first uses a move-until-touch controller to establish contact , followed by a joint-torque PD control. 
If contact forces exceed $10\,\mathrm{N}$ without displacement, the robot retracts and records the object as immovable. 


To handle hardware execution uncertainty and pose-estimation variability,  
\textsc{Precheck} applies a translation, orientation and joint configuration matching tolerances. 
Rotationally symmetric axes (e.g. mugs and markers) are ignored during orientation checks. 

The accompanying video demonstrates our framework across three two-object scenes in the Object-Movability and Pen-Functionality domains. 
In Object-Movability, when both the mug and can are movable, the robot verifies movability through interaction and transfers both objects to the mat.
When the mug is made immovable, the robot identifies this through the force-limited push, updates the state and remaining plan, and transfers the can while leaving the mug in place.
Finally, in Pen-Functionality, the before-and-after visual inspections during the writing tests allows the planner to distinguish between pens, placing the functional marker on the mat and returning the empty one on the table. 

\section{Conclusion and Discussions}


We presented a task and motion planning framework in which physical interaction reveals task-relevant properties.
Execution monitoring identifies invalidated actions and state mismatches, while recovery replanning reconnects execution to reusable portions of the plan.
Together, these mechanisms allow the robot to adapt to observed interaction outcomes and execution deviations while continuing to accomplish the task.


We adopt an explicitly optimistic viewpoint, planning under the assumption that objects are movable, pens functional, drawers openable and containers empty. 
Execution-time verification then triggers various replanning strategies based on observed outcomes.
In the Object-Movability and Drawer-Openability domains, negative outcomes render certain goals infeasible, prompting goal pruning, while managing post-interaction pose uncertainty. 
In the Pen-Functionality domain, goal pruning must additionally be paired with sequence of actions to return the non-functional pen back to the table. 
Conversely, in the Container-Emptiness domain, discovering a non-empty container does not invalidate the goal. 
Instead, it forces additional actions to pour the water out before placing it. 
These domains highlight how a unified optimistic framework balances goal revision, action insertion and execution recovery under uncertainty.








\balance
{\footnotesize
\bibliographystyle{ieeetr}
\bibliography{references}}

\end{document}